\documentclass[letterpaper]{article} % DO NOT CHANGE THIS
\usepackage{aaai2026}  % DO NOT CHANGE THIS
\usepackage{times}  % DO NOT CHANGE THIS
\usepackage{helvet}  % DO NOT CHANGE THIS
\usepackage{courier}  % DO NOT CHANGE THIS
\usepackage[hyphens]{url}  % DO NOT CHANGE THIS
\usepackage{graphicx} % DO NOT CHANGE THIS
\usepackage{natbib}  % DO NOT CHANGE THIS AND DO NOT ADD ANY OPTIONS TO IT
\usepackage{caption} % DO NOT CHANGE THIS AND DO NOT ADD ANY OPTIONS TO IT
\usepackage{algorithm}
\usepackage{algorithmic}
\usepackage{amsmath}
\usepackage{booktabs}
\usepackage{subcaption}
\usepackage{amssymb}

\usepackage{newfloat}
\usepackage{listings}
\DeclareCaptionStyle{ruled}{labelfont=normalfont,labelsep=colon,strut=off} % DO NOT CHANGE THIS
\floatstyle{ruled}
\newfloat{listing}{tb}{lst}{}
\floatname{listing}{Listing}
\title{CHDP: Cooperative Hybrid Diffusion Policies for Reinforcement Learning in Parameterized Action Space}
\author{
    Bingyi Liu\textsuperscript{\rm 1,2}, 
    Jinbo He\textsuperscript{\rm 1},
    Haiyong Shi\textsuperscript{\rm 1},
    Enshu Wang\textsuperscript{\rm 3}\thanks{Corresponding author.}, 
    Weizhen Han\textsuperscript{\rm 1}\footnotemark[1], 
    Jingxiang Hao\textsuperscript{\rm 3},
    Peixi Wang\textsuperscript{\rm 1},
    Zhuangzhuang Zhang\textsuperscript{\rm 4}
}
\affiliations{
    \textsuperscript{\rm 1}School of Computer Science and Artificial Intelligence, Wuhan University of Technology\\
    \textsuperscript{\rm 2}Hubei Key Laboratory of Transportation Internet of Things (Wuhan University of Technology)\\
    \textsuperscript{\rm 3}School of Cyber Science and Engineering, Wuhan University\\
    \textsuperscript{\rm 4}Department of Computer Science, City University of Hong Kong\\
    \{byliu, jinbohe, shihaiyong, hanweizhen, wpx20101220\}@whut.edu.cn, \{wanges17, haojingxiang\}@whu.edu.cn, zhuangzhuangzhang@cityu.edu.hk
}

\usepackage{bibentry}
\begin{document}

\maketitle

\begin{abstract}
Hybrid action space, which combines discrete choices and continuous parameters, is prevalent in domains such as robot control and game AI. However, efficiently modeling and optimizing hybrid discrete-continuous action space remains a fundamental challenge, mainly due to limited policy expressiveness and poor scalability in high-dimensional settings. 
To address this challenge, we view the hybrid action space problem as a fully cooperative game and propose a \textbf{Cooperative Hybrid Diffusion Policies (CHDP)} framework to solve it.
CHDP employs two cooperative agents that leverage a discrete and a continuous diffusion policy, respectively.
The continuous policy is conditioned on the discrete action's representation, explicitly modeling the dependency between them.
This cooperative design allows the diffusion policies to leverage their expressiveness to capture complex distributions in their respective action spaces.
To mitigate the update conflicts arising from simultaneous policy updates in this cooperative setting, we employ a sequential update scheme that fosters co-adaptation.
Moreover, to improve scalability when learning in high-dimensional discrete action space, we construct a codebook that embeds the action space into a low-dimensional latent space. 
This mapping enables the discrete policy to learn in a compact, structured space. 
Finally, we design a Q-function-based guidance mechanism to align the codebook's embeddings with the discrete policy's representation during training.
On challenging hybrid action benchmarks, CHDP outperforms the state-of-the-art method by up to $19.3\%$ in success rate.
\end{abstract}

% Uncomment the following to link to your code, datasets, an extended version or similar.
% You must keep this block between (not within) the abstract and the main body of the paper.
% \begin{links}
%     \link{Code}{https://aaai.org/example/code}
%     \link{Datasets}{https://aaai.org/example/datasets}
%     \link{Extended version}{https://aaai.org/example/extended-version}
% \end{links}

\section{Introduction}
Deep Reinforcement Learning (DRL) has excelled in domains with either purely discrete or continuous actions~\citep{duan2021distributional,duan2025distributional,liu2024multi}. 
However, extending these successes to real-world scenarios requires tackling hybrid action space. 
Such a space combines categorical choices (e.g., selecting a specific tool in robotic manipulation, choosing a movement mode in autonomous driving) with fine-grained continuous adjustments (e.g., regulating force magnitude, tuning speed parameters). 

Existing works~\citep{masson2016reinforcement,massaroli2020neural} face two primary challenges in tackling these hybrid action tasks. 
The first is that current policies lack the expressiveness to handle the multi-modality of many hybrid-action tasks.
This is because such multi-modality, where multiple action-pairs can be equally effective, conflicts with the unimodal architectures (e.g., Gaussian or deterministic) on which most methods are built~\citep{fan2019hybrid, HausknechtS15a}.
For example, a soccer goal can be scored with a left- or right-foot shot—--distinct actions, each with unique continuous parameters (e.g., force, angle)~\citep{masson2016reinforcement}.  
Consequently, such policies are forced to either average solutions into an ineffective compromise or collapse to a single mode, resulting in limited expressiveness and suboptimal performance~\citep{jainsampling,wang2024learning,huang2023reparameterized}.

Second, existing methods~\citep{xiong2018parametrized,fan2019hybrid} face a scalability limitation, as they lack a strategy to overcome the combinatorial explosion in high-dimensional hybrid action space~\citep{li2022hyar}. 
For instance, in non-prehensile manipulation, this challenge arises from selecting among numerous discrete contact points, each defined by its own set of continuous parameters~\citep{zhou2023hacman,le2025hydo}.
The difficulty of exploring such vast hybrid space leads to sample inefficiency, rendering most approaches impractical for these tasks~\citep{pmlr-v235-zhang24r}.
Addressing these dual challenges of policy expressiveness and limited scalability is therefore crucial for unlocking DRL's potential in complex hybrid-action environments.

To address these challenges, we propose \textbf{Cooperative Hybrid Diffusion Policies (CHDP)}, a multi-agent reinforcement learning (MARL) framework that views the hybrid-action problem as a fully cooperative game between two agents.
CHDP consists of two cooperative agents that leverage a discrete and a continuous diffusion policy, respectively.
These agents collaborate to generate a final hybrid action aimed at maximizing a shared objective. 
% CHDP consists of a discrete agent and a continuous agent, each empowered by its own diffusion policy.
This collaboration is implemented sequentially, where the continuous policy is conditioned on the discrete action representation to explicitly model the dependency between them.
This design leverages the expressiveness of diffusion policies to capture complex, multi-modal distributions across their respective action spaces.
However, this cooperative setting can lead to conflicts when policies are updated simultaneously, a challenge empirically observed in MARL~\citep{lowe2017multi,yu2022surprising}. Our framework employs a sequential policy update training scheme inspired by prior works~\citep{wang2023order,liu2024maximum,JMLR:v25:23-0488}.
This mechanism prevents the optimization of one policy from inadvertently impairing the other, thus fostering co-adaptation and allowing CHDP to solve complex hybrid-action tasks.
Moreover, to improve scalability when learning in high-dimensional discrete action space, we construct a codebook that embeds the high-dimensional discrete action space into a low-dimensional latent space.
The discrete policy generates a continuous latent vector, which is then quantized by mapping it to its nearest neighbor in the embedding codebook. 
This quantization process confines the discrete policy's learning to a compact, structured space, thereby mitigating the curse of dimensionality. 
Subsequently, the selected embedding serves as a semantic condition to guide the continuous policy. 
Finally, we design a shared, Q-function-based guidance mechanism to align the codebook's embeddings with the latent representations generated by the discrete policy during training. 

Our main contributions are summarized as follows:
\begin{itemize}
    \item We propose CHDP, a MARL framework that solves the hybrid action space problem by fully leveraging two cooperative agents with diffusion policies. This design unleashes the expressiveness of the diffusion policies to capture multi-modal distributions in hybrid action space.
    \item We propose two key mechanisms: a sequential update scheme to mitigate policy conflicts and a Q-guided codebook to address scalability limitations through a compact action representation.
    \item CHDP demonstrates state-of-the-art (SOTA) performance on challenging hybrid action benchmarks, improving upon the prior SOTA method by up to $19.3\%$.
\end{itemize}

\section{Related Work}

\textbf{The Challenge of Limited Scalability.}
While Deep Reinforcement Learning (DRL) has become the dominant approach for the hybrid action space problem~\citep{HausknechtS15a, fan2019hybrid}, its various architectural explorations have consistently faced challenges with scalability and stability.
Early attempts to homogenize hybrid action space, by either discretizing continuous actions or converting discrete ones into continuous, proved ineffective due to the curse of dimensionality and the creation of poorly structured policy functions, respectively~\citep{li2022hyar}.
Another line of research, centered on the Parameterized DQN (PDQN)~\citep{xiong2018parametrized} suffers from critical redundancy and scalability issues. HACMan~\citep{zhou2023hacman} employs a similar architecture for non-prehensile manipulation, inheriting similar challenges. 
Meanwhile, hierarchical approaches like the Hierarchical Hybrid Q-Network (HHQN)~\citep{fu2019deep} are plagued by severe non-stationarity~\citep{wang2021i2hrl}. 
Collectively, these works lack an architecture that can scale effectively in domains with hybrid action space.
% Unlike these approaches that struggle with scalability, CHDP employs a Q-guided codebook to build a compact, structured representation.

\textbf{The Limitation of Policy Expressiveness.}
Running parallel to the scalability issue is the limitation of policy expressiveness. This trade-off is evident in recent methods: while HyAR~\citep{li2022hyar} offers a scalable latent-space framework, its expressiveness is constrained by its underlying deterministic policy and is often hampered by significant training instability~\citep{liu2024hybrid}.
Conversely, HyDo~\citep{le2025hydo} employs an expressive continuous diffusion policy, but its underlying architecture inherits the scalability limitations of methods like PDQN. 
Other works have focused solely on enhancing expressiveness through methods like energy-based models~\citep{jainsampling}, reparameterization in RPG~\citep{huang2023reparameterized}, GMMs in LOM~\citep{wang2024learning}, and Diffusion Q-learning~\citep{wang2023diffusion}. 
However, these methods lack the necessary expressiveness to capture the multi-modality for both discrete and continuous action spaces.

In summary, the methods discussed above suffer from poor scalability and limited expressiveness. We propose CHDP, a novel framework that bridges this gap by combining the scalability of a Q-guided codebook with the rich expressiveness of diffusion policies.

\section{Preliminaries}
% \subsection{Hybrid Action MDP as a Cooperative Game}
\subsection{Parameterized Action MDP (PAMDP)}
In this paper, we distinguish between two types of timesteps. We use superscripts $i \in \{1, \dots, N\}$ to denote the diffusion process timestep, and subscripts $t \in \{1, \dots, T\}$ to denote the reinforcement learning trajectory timestep.

The problem of controlling systems with hybrid actions is formalized as a Parameterized Action Markov Decision Process (PAMDP)~\citep{HausknechtS15a}.
A PAMDP is defined by a tuple $(\mathcal{S}, \mathcal{A}, P, r, \gamma)$, consisting of a state space $\mathcal{S}$, a hybrid action space $\mathcal{A}$, a transition function $P: \mathcal{S} \times \mathcal{A} \to \mathcal{P}(\mathcal{S})$, a reward function $r: \mathcal{S} \times \mathcal{A} \to \mathbb{R}$, and a discount factor $\gamma \in [0, 1)$.
The key feature of a PAMDP lies in its hybrid action space, where an action $\mathbf{a} \in \mathcal{A}$ is a pair $\mathbf{a} = (a^d, a^c)$.
This pair comprises a discrete action $a^d \in \mathcal{A}_d$ and a vector of continuous parameters $a^c$ from the discrete action-dependent set $\mathcal{A}_c(a^d)$.

To solve the PAMDP, we approach it from a MARL perspective. 
We decompose the single agent's decision-making process into a cooperative task executed by two agents. 
A discrete agent, guided by its policy $\pi_{\theta_d}(e_t|{s}_t)$, first selects a representation $e_t$.
This representation is then used to derive both the discrete action $a_t^d$ and its corresponding embedding, $e_k$.
Subsequently, a continuous agent determines the corresponding parameters $a_t^c$ via its policy $\pi_{\theta_c}(a_t^c|{s}_t, e_k)$, which is conditioned on both the state and the discrete action's embedding.
Here, $\theta_d$ and $\theta_c$ are the learnable parameters of their respective policies.
The objective is to learn a joint policy $\pi = (\pi_{\theta_d}, \pi_{\theta_c})$ that maximizes the expected discounted return ${\mathbb{E}\left[\sum_{t=0}^{T} \gamma^t r({s}_t, \mathbf{a}_t)\right]}$.
% where $\tau = (\mathbf{s}_0, \mathbf{a}_0, \mathbf{s}_1, \mathbf{a}_1, \dots)$ is a trajectory sampled under policy $\pi$ over a time horizon $T$.
Correspondingly, the action-value (or Q-value) for the joint policy $\pi$ is defined as $Q^\pi_\phi({s}_t, e_t, a_t^c) = \mathbb{E}_{\pi}\left[\sum_{k=0}^{\infty} \gamma^k r({s}_{t+k}, \mathbf{a}_{t+k}) | {s}_t, e_t, a_t^c\right]$, where $\phi$ denotes the parameters of the Q-function.
\begin{figure*}[!t]

\centering

\includegraphics[width=0.98\linewidth]{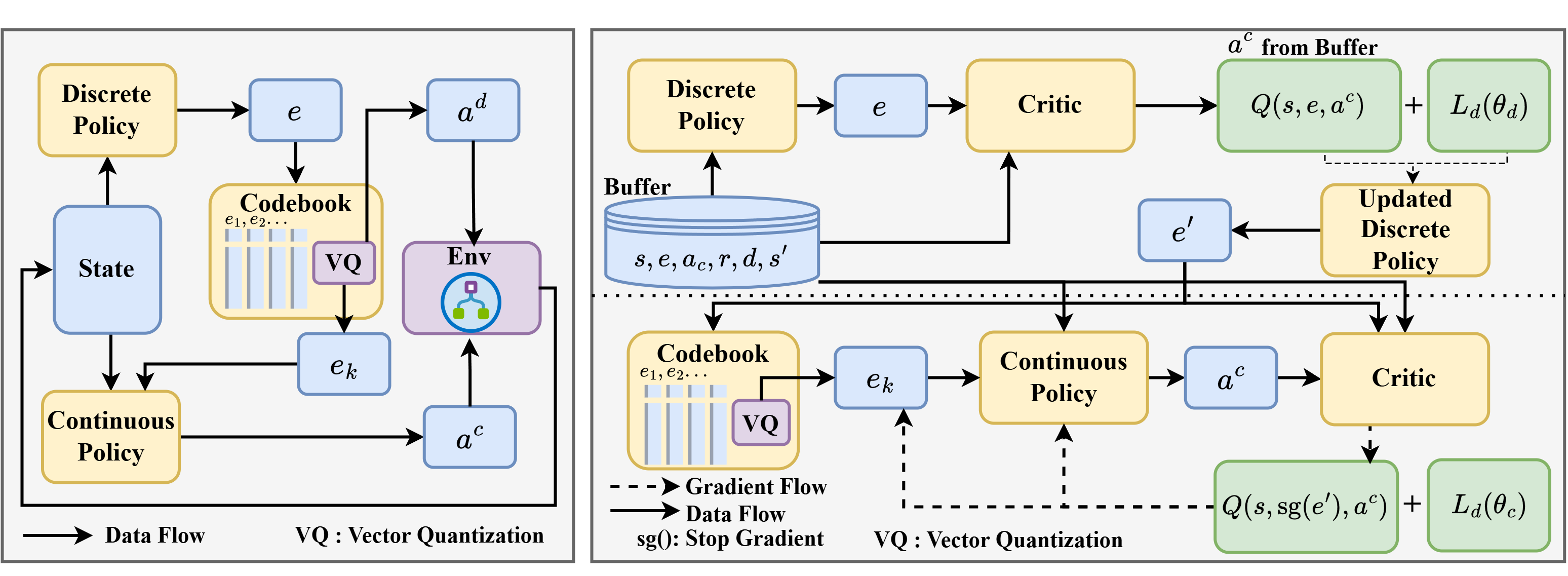}
\caption{An overview of our CHDP framework.
    {(Left) Inference:} The discrete policy's latent representation is quantized by a codebook to yield a discrete action and relevant codeword. This codeword conditions the continuous policy for generating the final continuous action.
    {(Right) Training:} Guided by a shared Q-function, the discrete policy is updated first. Subsequently, conditioned on the output of this updated discrete policy, the continuous policy and the codebook are jointly optimized.}
\label{fig:overview}
\end{figure*}

\subsection{Diffusion Models and DQL}

\textbf{Diffusion Models for Generative Learning.}
Our framework builds upon diffusion models~\citep{ho2020denoising, sohl2015deep, song2019generative}, a class of powerful generative models. A diffusion model learns to generate data by reversing a fixed {forward process} that gradually adds Gaussian noise to data $a_0$ over $N$ steps, transforming it into pure noise $a_N \sim \mathcal{N}(0, I)$.

The core of the model is to learn the {reverse process}. This is achieved by training a neural network, $\epsilon_\theta(a_i, s, i)$, to predict the noise that was added at each step $i$. Once trained, this network can be used to iteratively denoise a random noise vector to generate a new data sample $a_0$. Each step of this reverse process is performed as follows:
\begin{equation} 
    % \label{eq:denoising_step}
    a_{i-1} = \frac{1}{\sqrt{\alpha_i}} \left( a_i - \frac{1-\alpha_i}{\sqrt{1-\bar{\alpha}_i}} \epsilon_\theta(a_i, s, i) \right) + \sqrt{\beta_i}z,
\end{equation}
where $\alpha_i, \bar{\alpha}_i, \beta_i$ are hyperparameters from a predefined noise schedule, and $z \sim \mathcal{N}(0, I)$. 
The network $\epsilon_\theta$ is trained by the diffusion loss:
{\footnotesize
    \begin{equation} \label{eq:standard_diffusion_loss}
    \mathcal{L}_d(\theta) = \mathbb{E}_{s, a_0, \epsilon, i} \left[ \|\epsilon - \epsilon_\theta(\sqrt{\bar{\alpha}_i}a_0 + \sqrt{1-\bar{\alpha}_i}\epsilon, s, i)\|^2 \right].
\end{equation}
}

\textbf{Diffusion Q-Learning (DQL).}
Diffusion Q-Learning (DQL)~\citep{wang2023diffusion} adapts this generative framework to serve as an expressive policy in reinforcement learning. 
Instead of merely modeling the data distribution, a DQL policy must generate actions that maximize long-term returns. 
To achieve this, DQL augments the standard diffusion loss $\mathcal{L}_d(\theta)$ with a {policy improvement term}, $\mathcal{L}_q(\theta)$, which guides the policy towards high-value actions.

The overall training objective for a DQL policy is a weighted sum of these two components:
\begin{equation} \label{eq:DQL_loss}
    \mathcal{L}(\theta) = \mathcal{L}_d(\theta) + \alpha \cdot \mathcal{L}_q(\theta).
\end{equation}
The policy improvement term $\mathcal{L}_q(\theta)$ is defined as the negative expected Q-value of the generated actions:
\begin{equation} \label{eq:dql_q_loss}
    \mathcal{L}_q(\theta) = -\mathbb{E}_{s \sim \mathcal{D}, a_0 \sim \pi_\theta}[Q_\phi(s, a_0)].
\end{equation}
DQL sets $\alpha = \eta / \mathbb{E}_{s,a \sim \mathcal{D}}[|Q_\phi(s, a)|]$, where $\eta$ is a hyperparameter that balances the two loss terms.

\section{Cooperative Hybrid Diffusion Policies}

\subsection{Overview of the CHDP Framework}
The proposed CHDP, depicted in Figure~\ref{fig:overview}, features a novel cooperative framework with three core components. 
CHDP consists of two specialized agents: a discrete agent using policy $\pi_{\theta_d}$ for selecting discrete actions, and a continuous agent using policy $\pi_{\theta_c}$ for specifying the corresponding continuous parameters. 
In addition, a sequential update mechanism ensures co-adaptation by mitigating policy conflicts.
Inspired by VQ-VAE~\citep{van2017neural}, we construct a learnable Q-guided codebook, $\mathbf{E}_{\zeta} \in \mathbb{R}^{K \times {d}_{e}}$, which embeds the high-dimensional discrete action space into a low-dimensional latent space. 
Comprising $K$ learnable codewords (${e}_k \in \mathbb{R}^{d_e}$), where $K$ is the cardinality of the discrete action space, the codebook also serves as the semantic bridge between the discrete and continuous policies. 

At each timestep, the action generation process unfolds as a structured, sequential collaboration. 
Given the current state $s$, the discrete policy, $\pi_{\theta_d}$, first generates a latent representation $e$ by iteratively denoising an initial Gaussian noise vector $e_N \sim \mathcal{N}(0, I)$. 
Each sampling step is defined as:
\begin{equation}
    e_{i-1} = \frac{1}{\sqrt{\alpha_i}} \left( e_i - \frac{1-\alpha_i}{\sqrt{1-\bar{\alpha}_i}} \epsilon_{\theta_d}(e_i, s, i) \right) + \sqrt{\beta_i}z,
\end{equation}
where $z \sim \mathcal{N}(0, I)$ is a noise vector and $\epsilon_{\theta_d}$ denotes the noise prediction network for the discrete policy. 
The final output, $e = e_0$, then undergoes a Vector Quantization (VQ)~\citep{linde2003algorithm} step, where it is mapped to the nearest entry in the codebook. 
This yields the index $k$ of the nearest codeword, which is set as the discrete action ($a^d=k$), and the corresponding codeword vector $e_k$, which is retrieved to condition the continuous policy.
The specific mechanism for this codebook will be detailed in Section~\ref{sec:codebook}.

Subsequently, the continuous policy $\pi_{\theta_c}$ generates the continuous action $a^c$. Critically, its reverse diffusion process is conditioned on both the state $s$ and the codeword $e_k$ from the first stage:
{\footnotesize
\begin{equation}
    a_{i-1}^c = \frac{1}{\sqrt{\alpha_i}} \left( a_i^c - \frac{1-\alpha_i}{\sqrt{1-\bar{\alpha}_i}} \epsilon_{\theta_c}(a_i^c, s, e_k, i) \right) + \sqrt{\beta_i}z,
\end{equation}
}where $\epsilon_{\theta_c}$ is the noise prediction network. This directly infuses the semantic information of the discrete choice into the continuous action generation, ensuring the two components are coherent.

\subsection{Sequential Update Scheme}
CHDP employs a sequential update scheme for policy updates, a strategy inspired by Heterogeneous-Agent Reinforcement Learning (HARL)~\citep{liu2024maximum,JMLR:v25:23-0488}.
As illustrated in Figure~\ref{fig:overview}(Right), this scheme decomposes the joint optimization into a turn-based sequence. 

\textbf{Step 1: Discrete Policy Update.}
The training process begins with the discrete policy.
CHDP first updates the discrete policy $\pi_{\theta_d}$, which provides a fixed, stable target for the subsequent update of the continuous policy. 
This update is performed by minimizing the objective function $\mathcal{L}(\theta_d)$ with the goal of improving the generated latent representation $e \sim \pi_{\theta_d}(\cdot|s)$, which is defined as:
\begin{equation}
% \begin{split}
\mathcal{L}(\theta_d) = \mathcal{L}_d(\theta_d) - \alpha \cdot \mathbb{E}_{s,a^c \sim \mathcal{D}, e \sim \pi_{\theta_d}}[Q_{\phi}(s, e, a^c)],
% \end{split}
\label{eq:discrete_policy_loss_chdp_final}
\end{equation}
where $\mathcal{L}_d(\theta_d)$ is the standard diffusion loss from DQL, as defined in Eq.~\eqref{eq:standard_diffusion_loss}, which regularizes the policy using data from the replay buffer. 
% , which serves as a policy regularizer based on data from the replay buffer.
The second term, $\mathcal{L}_q(\theta_d)$, is the policy improvement objective that guides the latent representation $e$ towards higher Q-values. 
The continuous action $a^c$ is sampled from the replay buffer $\mathcal{D}$, and this fixed value serves as a stable target for evaluating the latent representation $e$.
% By updating the discrete policy first, this step provides a fixed, stable target for the subsequent optimization of the continuous policy.
This initial step provides a stable foundation for the subsequent optimization of the continuous policy and codebook.

\textbf{Step 2: Continuous Policy and Codebook Update.}
In this second step, the newly updated discrete policy, $\pi'_{\theta_d}$, generates a latent representation $e' \sim \pi'_{\theta_d}(\cdot|s)$. 
This vector is quantized to select a codeword $e_k$ from the codebook $\mathbf{E}_\zeta$, which in turn conditions the continuous policy to produce an action $a^c \sim \pi_{\theta_c}(\cdot|s, e_{a^d})$. 
The continuous policy parameters, $\theta_c$, and the codebook parameters, $\zeta$, are then jointly optimized by minimizing their shared objective $\mathcal{L}(\theta_c, \zeta)$:
\begin{equation}
    \mathcal{L}(\theta_c, \zeta) = \mathcal{L}_d(\theta_c) + \alpha \cdot \mathcal{L}_q(\theta_c, \zeta),
    \label{eq:continuous_policy_loss_chdp_final}
\end{equation}
where the first term, $\mathcal{L}_d(\theta_c)$, is a diffusion loss analogous to the one for the discrete policy, as defined in Eq.~\eqref{eq:standard_diffusion_loss}.

The second term, $\mathcal{L}_q(\theta_c, \zeta)$, is the policy improvement term that guides the continuous policy towards generating actions that yield higher Q-values. It is defined as:
\begin{equation}
    \mathcal{L}_q(\theta_c, \zeta) = -\mathbb{E}_{s \sim \mathcal{D}, e' \sim \pi'_{\theta_d}, a^c \sim \pi_{\theta_c}} [Q_{\phi}(s, \operatorname{sg}(e'), a^c)].
\end{equation}
The design of this objective is crucial for managing the gradient flow.
Gradients from the Q-function update the continuous policy $\pi_{\theta_c}$ based on the output of the updated discrete policy, rather than the outdated discrete action from the replay buffer.
Meanwhile, a stop-gradient operator $\operatorname{sg}(\cdot)$ on the discrete representation prevents any gradient from flowing back to the discrete policy $\pi'_{\theta_d}$, thereby resolving potential conflicts in the policy updates.

Concurrently, the critics are updated by minimizing the Mean-Squared Bellman Error (MSBE). 
To ensure stability, we adopt the Double Q-learning paradigm~\citep{van2016deep}. The loss for each critic $Q_{\phi_i}$ is:
{\footnotesize
    \begin{equation}
    \mathcal{L}(\phi_i) = \mathbb{E}_{({s}_t, e_t, a_t^c, r_t, {s}_{t+1}) \sim \mathcal{D}} \left[ \left( Q_{\phi_i}({s}_t, e_t, a_t^c) - y_t \right)^2 \right],
\label{eq:CHDP_critic_loss_overview}
\end{equation}}where the target value $y_t$ is computed using the next action $\mathbf{a}_{t+1}$, which is sampled from the target policies $\pi'_{\theta'_d}$ and $\pi'_{\theta'_c}$.
The target value is then defined as:
\begin{equation}
   y_t = r_t + \gamma \min_{j=1,2} Q'_{\phi'_j}({s}_{t+1}, e_{t+1}, a_{t+1}^c).
\end{equation}
\begin{figure}[!t]
    \centering
    \includegraphics[width=0.95\columnwidth]{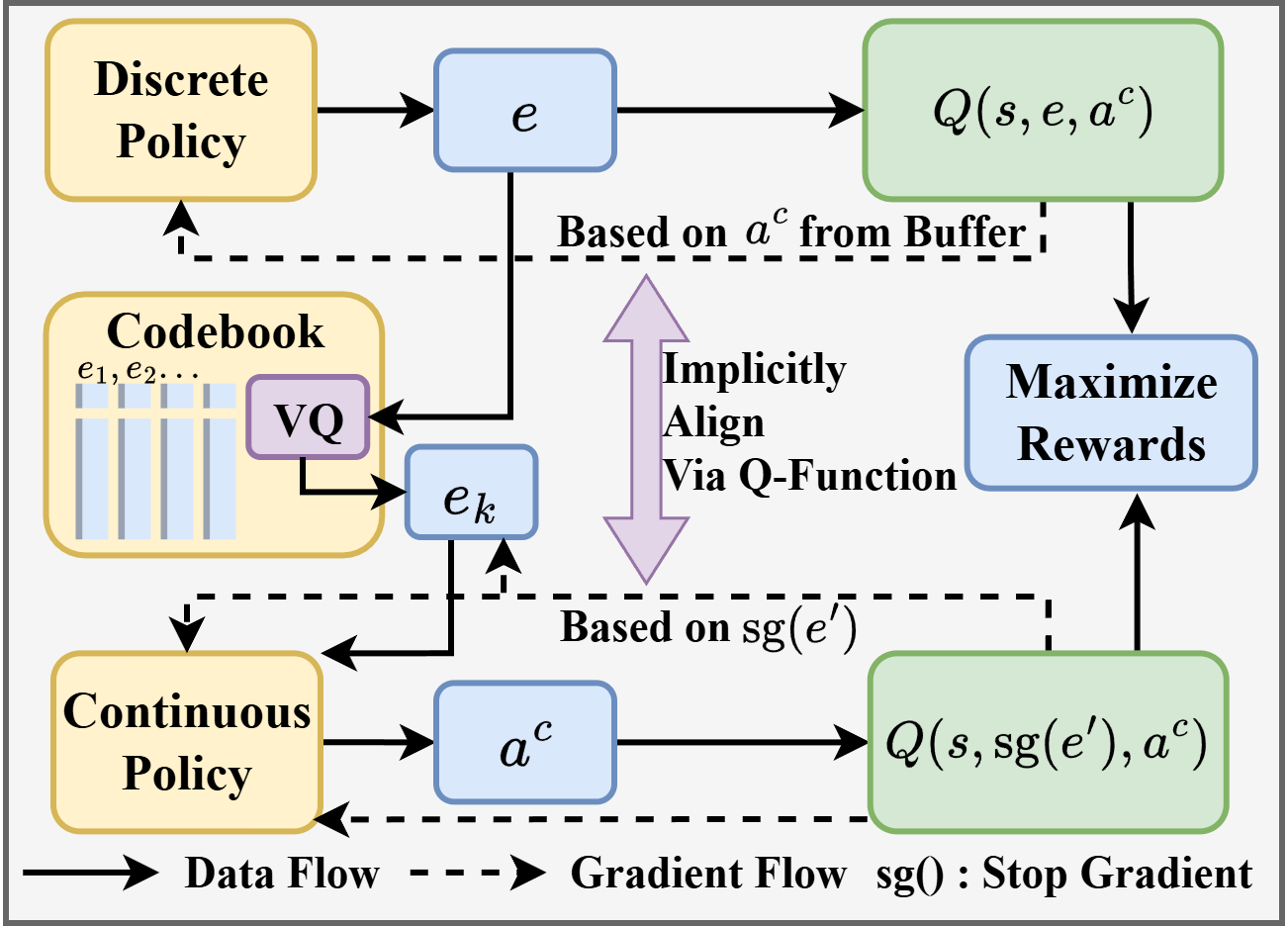}
    \caption{The process of aligning latent representations and codewords.}
    \label{fig:codebook}
\end{figure}

\subsection{Downstream Q-guided Codebook}
\label{sec:codebook}
CHDP constructs a learnable codebook $\mathbf{E}_{\zeta}$ to represent the discrete action space.
Instead of learning via reconstruction, the value of a codeword is measured by its functional impact on the downstream task, that is, its effectiveness in maximizing the Q-value. 
This is enforced by a deliberately asymmetric gradient flow, where the shared Q-function guides both the discrete policy's latent output and the selected codeword through distinct optimization pathways, forcing them to implicitly align.

The aligning process, as illustrated in Figure~\ref{fig:codebook}, begins with the discrete policy $\pi_{\theta_d}$ generating a latent representation $e \sim \pi_{\theta_d}(\cdot|s)$.
This vector is then quantized to its nearest codeword $e_k$ in the codebook $\mathbf{E}_{\zeta}$ via a VQ operation:
\begin{equation}
    % k = \text{argmin}_k \|e - e_k\|^2.
    k = \operatorname*{argmin}_k \|e - e_k\|^2.
\end{equation} 

The selected codeword $e_k$ is subsequently used to condition the continuous policy $\pi_{\theta_c}$, which generates the final continuous parameters. 

The codebook's parameters, $\zeta$, are optimized via gradients originating from the continuous policy's improvement term $\mathcal{L}_q(\theta_c, \zeta)$.
Since the continuous action $a^c$ is conditioned on the chosen codeword ${e}_{k}$, the final Q-value is an implicit function of this codeword. 
Consequently, during backpropagation, gradients from the Q-value propagate through the continuous action $a^c$ back to the specific codeword ${e}_{k}$ used in its generation. 
This downstream signal steers the codeword's embedding toward a representation that supports higher-value actions.

In contrast, the discrete policy, which generates the latent vector $e$, is optimized using the same Q-function, $Q(s, e, a^c)$, as defined in Eq.~\eqref{eq:discrete_policy_loss_chdp_final}. 
Crucially, for this update, the continuous action $a^c$ is a detached constant sampled from the replay buffer $\mathcal{D}$, severing the gradient path through it. 
In this way, both the codebook embeddings ${e}_{k}$ and the discrete policy's outputs ${e}$ are guided by the same ultimate measure of utility, the Q-value, forcing them to implicitly align within the shared latent space.
This training paradigm ensures the semantic meaning of each discrete action's embedding is forged by its downstream utility of maximizing the Q-value, rather than by reconstruction. This end-to-end process produces a task-aware, semantically potent representation of the discrete action space, ensuring a synergistic coupling between the policies. As a result, this integrated design provides a principled and scalable mechanism for representation learning that obviates the need for pre-training.

\begin{table*}[t]
\centering
\begin{tabular}{@{}l|c|cccc|c@{}}
\toprule
% \multirow{2}{*}
{ \textbf{ENV}} & \textbf{HPPO} & \textbf{PATD3} & \textbf{PDQN-TD3} & \textbf{HHQN-TD3} & \textbf{HyAR-TD3} &\textbf{CHDP}(ours) \\ 
% \cmidrule(l){2-7}
% & \textbf{PPO-based} & \multicolumn{4}{c|}{\textbf{TD3-based}} & \textbf{Diffusion-based} \\ 
\midrule
Goal & 0.0 $\pm$ 0.0 & 0.0 $\pm$ 0.0 & 71.4 $\pm$ 4.3 & 0.0 $\pm$ 0.0 & {77.3 $\pm$ 4.2} & \textbf{80.9 $\pm$ 4.9} \\

Hard Goal & 0.0 $\pm$ 0.0 & 43.0 $\pm$ 10.0 & 0.0 $\pm$ 0.0 & 1.2 $\pm$ 0.7 & {60.2 $\pm$ 5.0}  & \textbf{79.5 $\pm$ 5.0} \\

Platform & 66.3 $\pm$ 0.9 & 95.1 $\pm$ 3.6 & 96.7 $\pm$ 4.1 & 56.7 $\pm$ 29.4 & {96.6 $ \pm$ 2.2} & \textbf{99.7 $\pm$ 0.2} \\

Catch Point & 55.7 $\pm$ 5.2 & 86.7 $\pm$ 8.3 & 89.8 $\pm$ 3.2 & 23.7 $\pm$ 5.5 & 86.6 $\pm$ 0.9 & \textbf{93.8 $\pm$ 0.6} \\

Hard Move (n=4) & 3.3 $\pm$ 1.3 & 63.9 $\pm$ 20.5 & 79.7 $\pm$ 4.8 & 81.8 $\pm$ 3.5 & 91.4 $\pm$ 2.4 & \textbf{94.2 $\pm$ 1.7} \\

Hard Move (n=6) & 2.5 $\pm$ 0.4 & 9.8 $\pm$ 5.6 & 31.1 $\pm$ 8.1 & 47.1 $\pm$ 27.6  & {92.3 $\pm$ 0.6} & \textbf{93.9 $\pm$ 1.0} \\

Hard Move (n=8) & 2.3 $\pm$ 0.9 & 4.6 $\pm$ 2.1 & 6.6 $\pm$ 2.5 & 18.8 $\pm$ 8.9 & {88.3 $\pm$ 1.9} & \textbf{90.6 $\pm$ 2.2} \\

Hard Move (n=10) & 3.4 $\pm$ 0.5 & 10.3 $\pm$ 2.0 & 3.3 $\pm$ 0.5 & 11.3 $\pm$ 6.3 & 69.0 $\pm$ 5.6 & \textbf{79.8 $\pm$ 5.4} \\
\bottomrule
\end{tabular}
\caption{Mean success rates ($\pm$ one standard deviation) over 5 independent trials, where the result for each trial is itself the average of the success rates from its final 5 evaluations.}
\label{tab:results_wide}
\end{table*}

\section{Experiments}
We evaluate CHDP on various hybrid action environments against representative prior algorithms. Then, a detailed ablation study is conducted to verify the contribution of each component in CHDP. Moreover, we provide a qualitative experiment to show the expressiveness of the diffusion policy.

\begin{figure}[t]
    \centering
    \begin{subfigure}[t]{0.45\columnwidth}
        \centering
        \includegraphics[width=\linewidth, trim=0 50 0 50, clip]{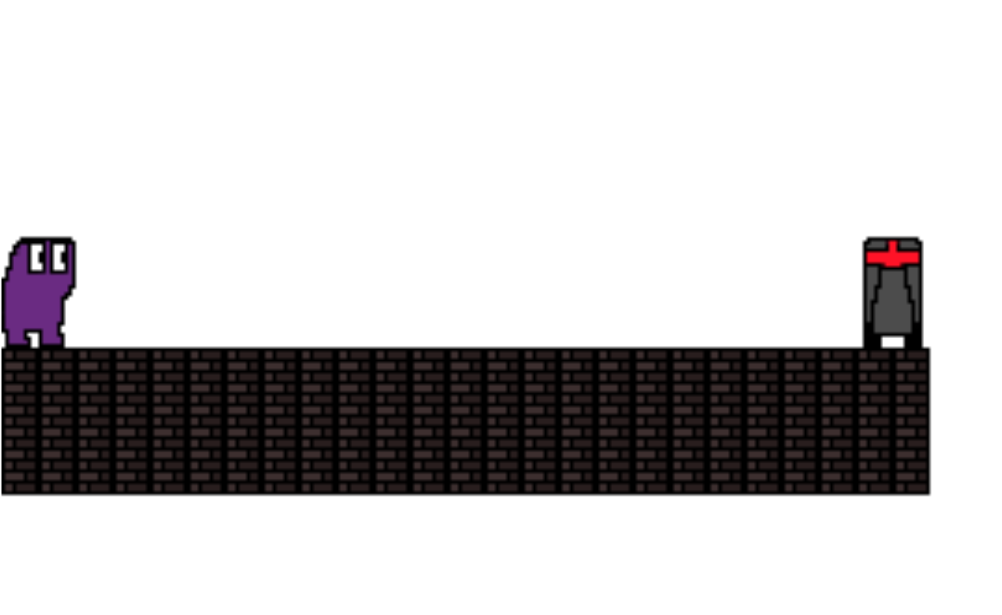}
        \caption{Platform}
    \end{subfigure}
    \hfill
    \begin{subfigure}[t]{0.45\columnwidth}
        \centering
        \includegraphics[width=\linewidth]{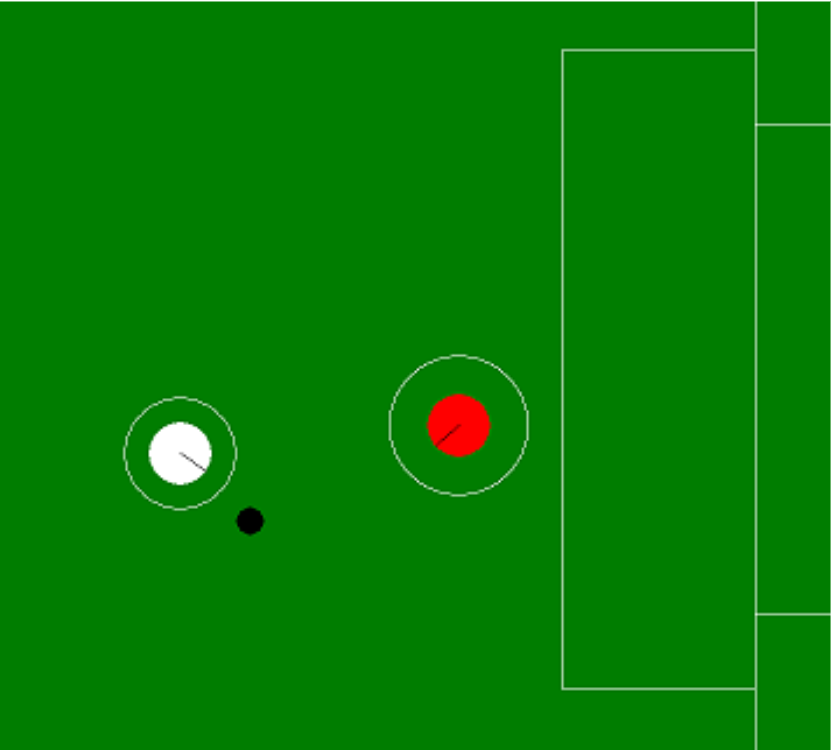}
        \caption{Goal}
    \end{subfigure}

    % --- 第二行 ---
    \begin{subfigure}[t]{0.45\columnwidth}
        \centering
        \includegraphics[width=\linewidth]{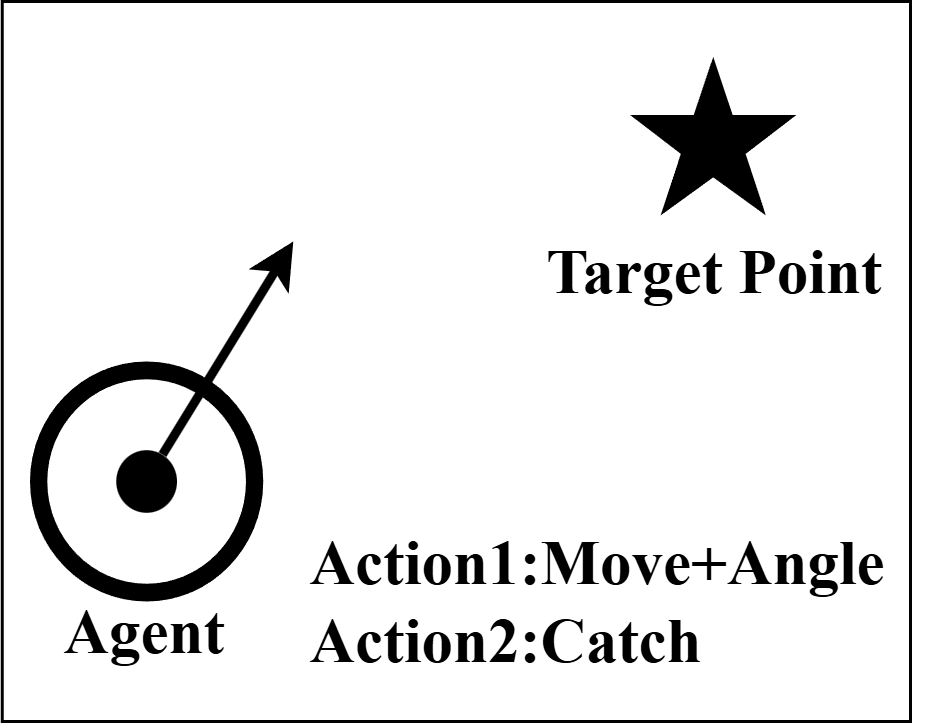}
        \caption{Catch Point}
    \end{subfigure}
    \hfill
    \begin{subfigure}[t]{0.45\columnwidth}
        \centering
        \includegraphics[width=\linewidth]{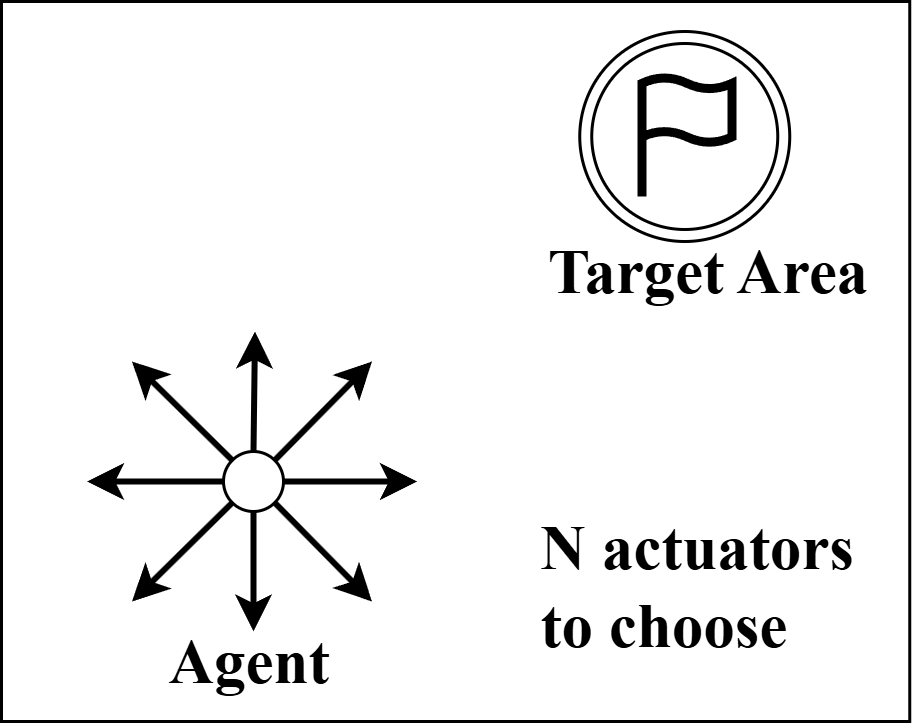}
        \caption{Hard Move}
    \end{subfigure}
    \caption{Visualizations of the tested environments.}
    \label{fig:environments}
\end{figure}

\textbf{Benchmarks:}
We evaluated CHDP on eight standard Parameterized Action MDP (PAMDP) benchmarks: Platform and Goal~\citep{masson2016reinforcement}, Catch Point~\citep{fan2019hybrid}, Hard Goal, and four variations of Hard Move~\citep{li2022hyar}. 
These environments are identical to those used to evaluate HyAR, which represents the SOTA algorithm for these tasks.
Visualizations of these environments are shown in Figure~\ref{fig:environments}. 

\textbf{Baselines:}
We evaluate CHDP against several SOTA baselines for hybrid action space. 
Our primary comparison is against HyAR~\citep{li2022hyar}, the current leading method. 
To ensure a fair and direct comparison, we adopt the same set of TD3-based baselines used in HyAR. 
These baselines extend prior methods using the Twin Delayed Deep Deterministic Policy Gradient (TD3) algorithm~\citep{fujimoto2018Addressing} and include: PDQN-TD3 (an extension of PDQN~\citep{xiong2018parametrized}), PA-TD3 (from PADDPG~\citep{HausknechtS15a}), and HHQN-TD3 (from HHQN~\citep{fu2019deep}). 
Additionally, we include the PPO-based HPPO~\citep{fan2019hybrid} for broader coverage. 

\subsection{Performance Evaluation}
\begin{figure*}[t]
\centering
\includegraphics[width=0.99\linewidth]{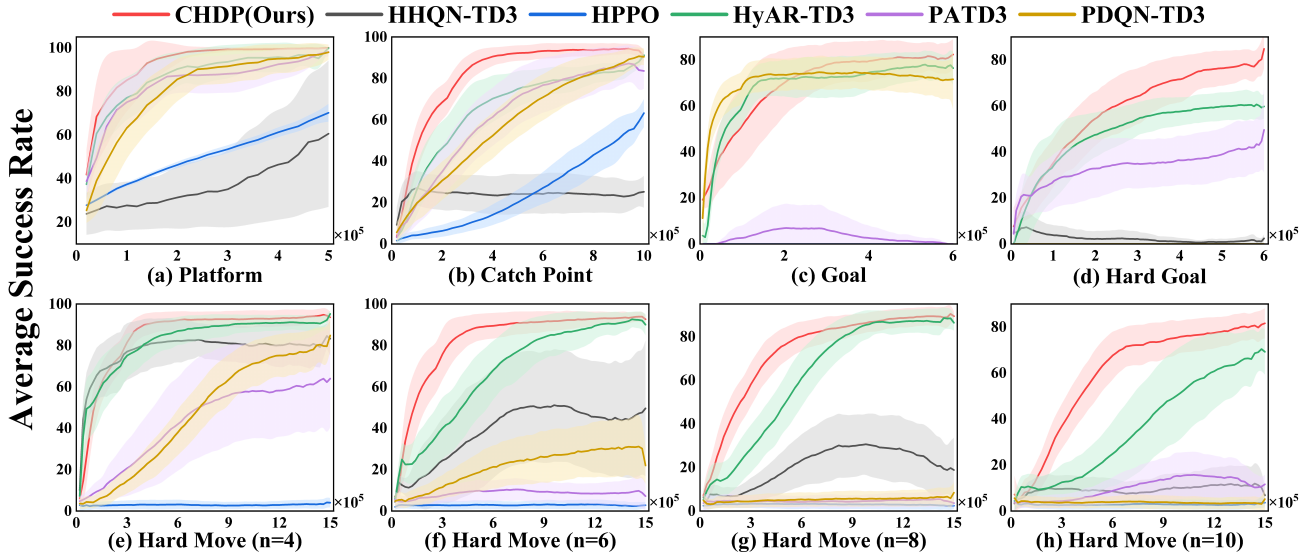}
\caption{Comparisons of algorithms in different environments. The x- and y-axis represent the environment steps ($\times 10^5$) and the average success rate, respectively. The solid curve and the shaded region denote the mean and standard deviation over 5 independent runs. All curves are smoothed for visual clarity.}
\label{fig:performance}
\end{figure*}
We conducted a comprehensive empirical study to evaluate the performance of our proposed method, CHDP, against a suite of strong baseline algorithms across eight standard PAMDP benchmarks. The mean success rates at the end of training are presented in Table~\ref{tab:results_wide}, and the learning dynamics are shown in Figure~\ref{fig:performance}. The results clearly demonstrate that CHDP achieves SOTA performance, consistently outperforming all baselines by showcasing superior policy expressiveness, scalability, and sample efficiency.

\textbf{Analysis of Policy Expressiveness.}
CHDP demonstrates significant performance gains in hybrid action environments that demand highly expressive policies to handle multi-modal action distributions. 
This is most evident in the \texttt{Hard Goal} task, where CHDP achieves a success rate of $79.5\%$, substantially surpassing the strongest baseline, HyAR-TD3, which scored only $60.2\%$.
This advantage extends across other challenging tasks. 
As shown in Table~\ref{tab:results_wide}, CHDP also secures SOTA performance in the \texttt{Platform} ($99.7\%$) and \texttt{Catch Point} ($93.8\%$) environments. 
% This consistent, top-tier performance across multiple distinct benchmarks underscores the benefit of CHDP's diffusion-based policies in capturing the complex, multi-modal strategies required for success.
This underscores the advantage of CHDP's diffusion-based policy in capturing complex, multi-modal strategies.

\textbf{Analysis of Scalability in High-Dimensional Space.}
The \texttt{Hard Move} suite of tasks is designed to test scalability as the size of the discrete action space increases.
In these environments, a discrete action involves selecting an on/off state for each of $n$ actuators, creating a combinatorial space of size $2^n$. Each of these discrete choices is paired with its own set of continuous parameters (e.g., for $n=4$, there are $2^4=16$ discrete actions, each with its own corresponding set of continuous parameters).
As shown in Table~\ref{tab:results_wide}, CHDP consistently outperforms all competing methods across all difficulty levels, from $n=4$ to $n=10$. 
Notably, CHDP maintains a success rate exceeding $90\%$ for $n$ up to 8 ($2^8=256$ options), a scale where other methods degrade significantly. This validates the excellent scalability of our Q-guided codebook approach.
% Notably, CHDP maintains a high success rate above $90\%$ for $n$ up to 8, demonstrating high performance even as the discrete action space expands to $2^8=256$ options, a domain where other methods degrade significantly. 
% This confirms the excellent scalability of our Q-guided codebook approach.

\textbf{Analysis of Convergence and Sample Efficiency.}
The learning curves in Figure~\ref{fig:performance} demonstrate CHDP's superior sample efficiency. 
Compared to all baselines, CHDP's learning curve (red) consistently exhibits a steeper ascent, indicating that it learns effective policies with substantially fewer environmental interactions. 
This advantage is particularly pronounced in \texttt{Platform} (a), \texttt{Catch Point} (b), and the \texttt{Hard Move} suite (e-h), where CHDP not only converges to a higher final performance but also does so more rapidly.
Such high sample efficiency not only validates the effectiveness of our approach but is also a crucial factor for its practical viability in complex, resource-intensive tasks.
% This sample efficiency confirms CHDP's effectiveness and ensures its viability for complex applications.
% This sample efficiency confirms CHDP's effectiveness and ensures its viability for complex applications.

\subsection{Ablation Study}
To quantitatively validate our design, we conducted an ablation study to isolate the roles of our components: the {expressive diffusion policies}, the Q-guided codebook, and the sequential update scheme. 
The results on the \texttt{Hard Goal} and \texttt{Hard Move ($n=6$)} environments are presented in Table~\ref{tab:ablation}. 
They demonstrate that each component is indispensable, addressing a distinct challenge within the framework.
% Results show that each component is indispensable, addressing a distinct challenge within the framework.
% We provide further ablations and an analysis of training time in the Appendix.
% Our ablation studies on hyperparameters and training time analysis, detailed in the Appendix, show that a low $N$ (e.g., 5) degrades performance, while an intermediate $d_e$ effectively balances expressiveness and optimization difficulty.

\begin{table}[htbp]
\centering
\begin{tabular}{lcc}
\toprule
\textbf{Method} & \textbf{Hard Goal} & \textbf{Hard Move} \\
\midrule
\textbf{CHDP (Full Model)} & \textbf{81.8 $\pm$ 3.5} & \textbf{94.4 $\pm$ 1.2} \\
\midrule
w/o Diffusion Policy & 49.8 $\pm$ 26.9 & 88.2 $\pm$ 5.6 \\
w/o Codebook & 73.3 $\pm$ 7.8 & 56.1 $\pm$ 12.5 \\
w/o Sequential Update & 53.5 $\pm$ 14.5 & 92.3 $\pm$ 4.5 \\
\midrule
w/o Both & 52.4 $\pm$ 13.7 & 54.6 $\pm$ 9.6 \\
\bottomrule
\end{tabular}
\caption{
Ablation of CHDP components. 
Values are mean success rates ($\pm$ std.) over 5 independent runs. 
The score for each run is the average of its final 5 evaluations.
The first row uses default parameters: Diffusion steps $N=15$, $\eta=10$ and $d_e=8$, distinct from the configuration in Table~\ref{tab:results_wide}.
The `w/o Both` variant removes both the Codebook and the Sequential Update scheme.
Best results are in \textbf{bold}.
}
\label{tab:ablation}
\end{table}
\textbf{The Ablation of Diffusion Policy.} To verify our central hypothesis that policy expressivity is crucial, we created a variant where the diffusion policies were replaced by deterministic ones.
This change led to a performance drop on both \texttt{Hard Goal} and \texttt{Hard Move}. Although this deterministic variant still achieved a 49.8\% success rate in \texttt{Hard Goal}—outperforming several baselines—this result is far from SOTA.
This starkly demonstrates that even with our cooperative framework, a policy constrained by expressiveness is insufficient.
The success of the full CHDP model is therefore directly attributable to the diffusion policy's ability to capture complex, multi-modal action distributions, validating our primary motivation.

\textbf{The Ablation of Codebook.}
The contribution of the Q-guided codebook is most evident in environments with a high-dimensional discrete action space. As illustrated in the \texttt{Hard Move ($n=6$)} environment, which features $2^6$ discrete actions, replacing the codebook with a simple \texttt{argmax} selection over raw outputs caused performance to collapse catastrophically from 94.4\% to 56.1\%. This result confirms the codebook's critical role in ensuring scalability. By embedding the high-dimensional space into a compact latent space, the codebook provides a structured, low-dimensional representation for the discrete policy to learn in.

\textbf{The Ablation of Sequential Update.}
To evaluate our sequential update scheme's role in mitigating optimization conflicts, we tested an ablation variant that updates both policies concurrently using data from the replay buffer, a mechanism identical to MADDPG~\citep{lowe2017multi}.
The concurrent update is detrimental in the more complex \texttt{Hard Goal} environment, where performance drops from 81.8\% to 53.5\%. 
We hypothesize this is because this task's challenge is policy coordination, making the prevention of conflicting updates critical. 
In contrast, the performance drop is minimal in the \texttt{Hard Move} environment (from 94.4\% to 92.3\%), where the main bottleneck is its high-dimensional action space.
This confirms that our sequential update scheme is a vital component for ensuring co-adaptation, especially in coordination-intensive tasks.

\textbf{The Ablation of Both.}
The `w/o Both` variant, which lacks both the codebook and the sequential update mechanism, exhibits a large performance degradation in both environments (52.4\% and 54.6\%).
This result provides evidence that CHDP's success stems from a {synergistic interplay} between its components, confirming that both the codebook and sequential update are indispensable.

\subsection{Qualitative Analysis of Expressive Policy}
\begin{table}[htbp]
\centering

\begin{tabular}{@{}l c r@{}}
\toprule
\textbf{Base Direction} & \textbf{Frequency} & \textbf{Continuous Action} \\
\midrule
\multicolumn{3}{l}{\textbf{CHDP (Ours)}} \\
\quad $(-0.5, -0.866)$ & 79.0\% & $-0.897 \pm 0.089$ \\
\quad $(0.0, -0.866)$  & 17.0\% & $-0.703 \pm 0.309$ \\
\quad $(0.75, 0.443)$  & 4.0\%  & $0.945 \pm 0.070$ \\
\midrule
\multicolumn{3}{l}{\textbf{HyAR}} \\
\quad $(0.75, 0.433)$ & 100.00\% & $0.445 \pm 0.000$ \\
\bottomrule
\end{tabular}
\caption{Action distribution statistics across 100 trials.}
\label{tab:expressivity_analysis}
\end{table}
To assess our policy's multi-modality, we designed a targeted experiment. 
In the \texttt{Hard Move (n=6)} environment, we placed the target area just above the agent's fixed starting position, which renders the task solvable in a single step. This setting isolates our analysis from the confounding effects of a multi-step trajectory. 
Our analysis then hinges on two key metrics: the \texttt{Base Direction}, representing the resultant force vector from one of $2^6=64$ possible actuator combinations, and the \texttt{Continuous Action}, a scalar value to control the intensity of the final acceleration.

Our analysis compares CHDP against the deterministic baseline HyAR. The Gaussian baseline, HPPO, is not included as it failed to solve the task.
As detailed in Table~\ref{tab:expressivity_analysis}, {HyAR}'s policy collapses into a single, rigid mode, consistently selecting the same actuator combination.
This results in a fixed \texttt{Base Direction} and an unvarying \texttt{Continuous Action} with zero variance, demonstrating its inability to capture the task's multi-modality. 
In contrast, CHDP leverages its expressiveness to discover a multi-modal policy, employing at least three distinct actuator combinations with frequencies of 79.0\%, 17.0\%, and 4.0\%. 
Most notably, CHDP's policy demonstrates a sophisticated understanding of the task's dynamics by mastering different strategies. 
For instance, its primary approach (79.0\% frequency) involves pairing a southwest-pointing \texttt{Base Direction}, (-0.5, -0.866), with a strong negative \texttt{Continuous Action} to effectively invert the force vector and reach the objective. 
In stark contrast, it simultaneously discovers a more intuitive, northeast-pointing strategy, (0.75, 0.443), coupled with a positive \texttt{Continuous Action}. 
This ability to identify and exploit multiple, non-intuitive, and even opposing pathways provides compelling evidence of CHDP's superior expressiveness.
\section{Conclusion}
In this paper, we have proposed CHDP, a novel framework that addresses hybrid action space problems using two cooperating agents, each driven by a diffusion policy. The cooperative design lies in modeling the complex dependencies between actions by conditioning the continuous policy on the discrete action's representation. To ensure both stability and efficiency, CHDP further incorporates a sequential update scheme to mitigate update conflicts and a Q-guided codebook to create a compact yet expressive representation for the discrete action space.
Experimental results demonstrate that CHDP achieves SOTA performance on challenging benchmarks, outperforming the prior leading method by up to $19.3\%$. 
We further designed a qualitative experiment to evaluate the framework's expressiveness and scalability. 
In the \texttt{Hard Move ($n=6$)} benchmark, CHDP successfully learned at least three distinct strategies, despite its large discrete action space of $64$ ($2^6$) options.

\section*{Acknowledgements}
This work was supported in part by the National Natural Science Foundation of China under Grant 62272357 and 62302326; in part by Wuhan Science and Technology Joint Project for Building a Strong Transportation Country under Grant 2024-2-7; in part by The State Key Laboratory of Integrated Services Networks under Grant ISN25-09, and in part by the Wuhan Science and Technology Project for Key Research and Development \text{No.}~2024050702030090.

\bibliography{aaai2026}

\end{document}